\documentclass[11pt,a4paper]{article}
\usepackage[hyperref]{acl2021}
\usepackage{times}
\usepackage{latexsym}

\usepackage{microtype}

\usepackage{url}
\usepackage{bm}
\usepackage[utf8]{inputenc} 
\usepackage{booktabs}       
\usepackage{amsfonts}       
\usepackage{nicefrac}       

\usepackage{bbm}
\usepackage{graphicx, calc}
\usepackage{enumitem}
\usepackage{booktabs}
\usepackage{multirow}
\usepackage{graphicx}
\usepackage{caption}
\usepackage{subcaption}

\usepackage{xspace}
\usepackage{amssymb}

\usepackage{amssymb}
\usepackage{pifont}
\newcommand{\cmark}{\ding{51}}%
\newcommand{\xmark}{\ding{55}}%

\usepackage{arydshln}

\usepackage{amsmath}

\usepackage[normalem]{ulem} 

\newcommand{\Table}[1]{Table~\ref{#1}}

\newcommand{\Figure}[1]{Fig.~\ref{#1}}

\newcommand{\Section}[1]{Sec.~\ref{#1}}
\newcommand{\ignore}[1]{}

\newcommand\ie{\emph{i.e.}}
\newcommand\eg{\emph{e.g.}}

\newcommand\vilbert{ViLBERT\xspace}

\newcommand{\datasetName}[1]{SVO-Probes}

\newcommand{\mmt}{MMT\xspace}         
\newcommand{\mmtSS}{Merged--MMT\xspace}   
\newcommand{\mmtLQ}{Lang--MMT\xspace}   
\newcommand{\mmtIQ}{Image--MMT\xspace}   
\newcommand{\smt}{SMT\xspace}   
\newcommand{\mmtNoMLM}{No-MLM--MMT\xspace}   
\newcommand{\mmtNoMRM}{No-MRM--MMT\xspace}   

\newcommand\zsflickr{\emph{ZS Flickr}\xspace} %

\title{Do Image-Language Transformers Understand Verbs?}

\aclfinalcopy

\title{Probing Image--Language Transformers for Verb Understanding}

 \author{Lisa Anne Hendricks ~ Aida Nematzadeh \\
 DeepMind
 \\ \tt \footnotesize{\{lmh, nematzadeh\} @deepmind.com}}

\begin{document}

\maketitle

\begin{abstract}
Multimodal image--language transformers have achieved impressive results on a
   variety of tasks that rely on fine-tuning (\eg, visual question answering
   and image retrieval). We are interested in shedding light on the quality of
   their pretrained representations -- in particular, if these models can
   distinguish different types of verbs or if they rely solely on nouns in a
   given sentence. 
To do so, we collect a dataset of image--sentence pairs (in English) consisting
   of 421 verbs that are either visual or commonly found in the pretraining
   data (\ie, the Conceptual Captions dataset).
We use this dataset to evaluate pretrained image--language transformers and
   find that they fail more in situations that require verb understanding
   compared to other parts of speech. We also investigate what category of
   verbs are particularly challenging. 

\end{abstract}

\section{Evaluating Verb Understanding}

The success of image--language models in real-world applications relies on
their ability to relate different aspects of language (such as verbs or
objects) to images, which we refer to as multimodal understanding.  For
example, an image-retrieval model needs to distinguish between ``eating an
apple'' and ``cutting an apple'' and a captioning model must accurately
describe the actions in a scene.

Previous work shows that image--language benchmarks do not always fully measure
such multimodal understanding: object retrieval models fail to account for
linguistic structure \cite{akula2020words}, visual question answering (VQA)
models overly rely on language priors
\citep[][]{goyal2017making,agrawal2018don}, and captioning metrics do not
always measure if captions ``hallucinate'' objects in an image
\citep{rohrbach2018object}.  Inspired by this, prior work introduced tasks to
specifically examine whether models can relate objects to images
\citep{shekhar-etal-2017-foil} or classify frequent interactions associated
with objects \citep{chao:iccv2015}.
However, both these datasets are limited to the 80 objects in the MSCOCO
detection challenge \citep{lin2014microsoft}.

\begin{figure*}[t]
\centering{
\includegraphics[width=.93\linewidth]{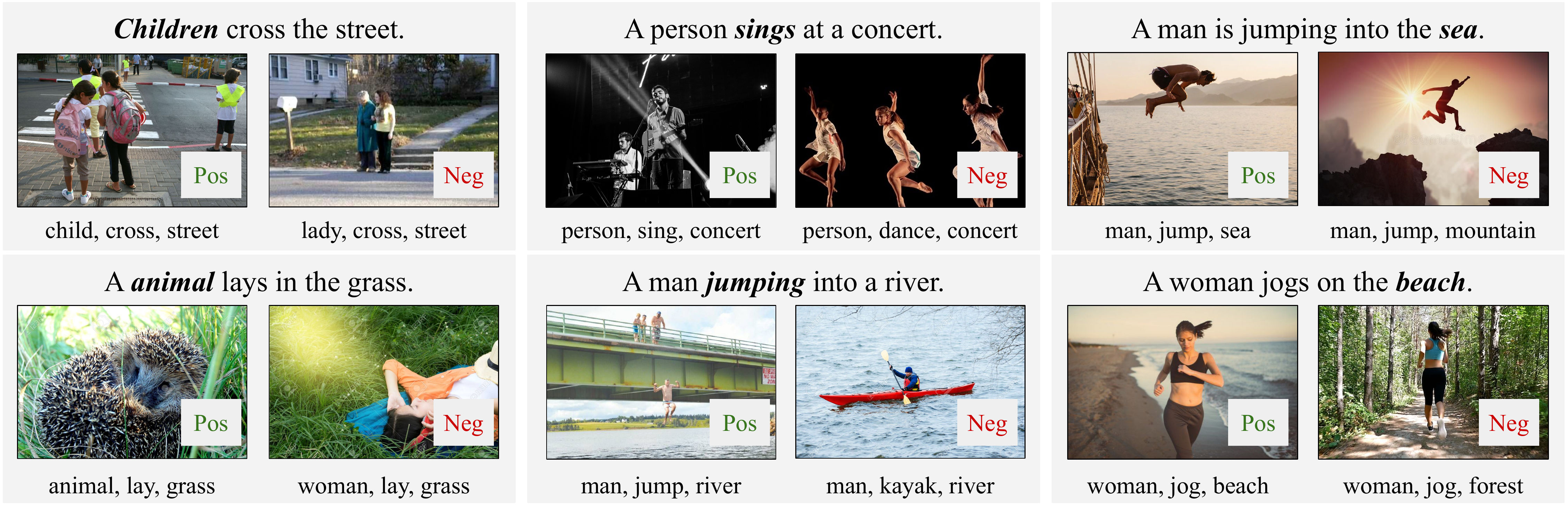}
\caption{Examples from \datasetName{}.  Images on the left and right show positive and negative image examples for each sentence.  Below each image is the $\langle$subject, verb, object$\rangle$ triplet corresponding to the image.}
\label{fig:qual-examples}
}
\end{figure*}

To address this gap, we design a benchmark focused on verbs called
\datasetName{} for examining \textbf{s}ubject,  \textbf{v}erb, \textbf{o}bject
triplets; more specifically, we collect a set of image--sentence pairs (in
English) where each pair is annotated with whether the sentence corresponds to
the image or not.
As shown in \Figure{fig:qual-examples}, for a given sentence, in addition to a
\emph{positive} image that matches the sentence, our dataset includes
controlled \emph{negative} images that do not correspond to specific aspects of
the sentence (\ie, subject, verb, and object).  These controlled examples
enable us to probe models for their understanding of verbs as well as subjects
and objects. Our dataset consists of $421$ verbs and includes over $48,000$
image--sentence pairs. 

We use our benchmark to evaluate the recent family of multimodal
(image--language) transformers that have shown impressive results on benchmarks
like VQA and image retrieval
\cite[][]{lu2019vilbert,chen2019uniter,tan2019lxmert,li2020oscar,Li2020unicoder,huang2020pixel}.
Our goal is to investigate if the good performance of these models is due to
learned representations that successfully relate different aspects of language
to images. More specifically, we evaluate a few architectural variations of
these models in a zero-shot way by using the pretrained models to classify if
image--sentence pairs from \datasetName{} match.

Our results show that the performance of all  evaluated models is worst on
verbs, with subjects being easier than verbs but harder than objects. We  find
that this observation does not depend on the frequency of test examples in
pretraining data. Moreover, it is considerably harder for all models to
correctly classify image--sentence pairs that do not match; the image--language
transformers overpredict that sentences corresponds to images.

Additionally, we compare an image--language transformer pretrained on a large
automatically-curated dataset \citep[\ie, Conceptual
Captions,][]{sharma2018conceptual} with one pretrained on the smaller but
manually-annotated MSCOCO \citep{chen2015microsoft}. Conceptual Captions is
more noisy than MSCOCO in that its sentences do not necessarily correspond to
its images. Interestingly, we observe that the model pretrained on MSCOCO
performs better. 
This result shows that the image--language transformers are not robust to
dataset noise as they learn to predict that somewhat-related image--sentence
pairs correspond to each other.

Despite their good performance on downstream tasks, image--language
transformers fail on our task that requires multimodal understanding {since
they cannot distinguish between finer-grained differences between images.  Our
results highlight that there is still considerable progress to be made when
training multimodal representations, and that verbs in particular are an
interesting challenge in image--language representation learning.}

\section{Related Work}
\label{sec:related_work}

Image--language transformers build on the transformer architecture
\citep[][]{vaswani2017attention} by incorporating additional loss functions (to
learn image features and align image and language modalities), using
self-attention to combine modalities, and training on paired image--text data
\citep{lu2019vilbert,chen2019uniter,tan2019lxmert,li2020oscar,Li2020unicoder,huang2020pixel}. 
The impressive performance of these models on many image--language benchmarks
has inspired recent work that studies different architectural choices made in
these models \citep{cao2020behind, hendricks2021decoupling}.

Compared to previous image--language models, multimodal transformers both use a
new architecture and are frequently trained on a much larger dataset -- the
Conceptual Captions dataset consisting of ~$3m$ image--text pairs
\citep{sharma2018conceptual}.  \citet{singh2020we} show that on fine-tuned
tasks, the performance of multimodal transformers
\citep[\ie,][]{lu2019vilbert,li2019visualbert} are less sensitive to  dataset
size; the domain match between pretraining and fine-tuning datasets is more
important.

\paragraph{Datasets.} Our proposed dataset is most similar to the FOIL
benchmark \citep{shekhar-etal-2017-foil} which tests if image--language models
can differentiate between sentences that vary with respect to only one noun.
FOIL consists of $64,300$ images from MSCOCO  \citep{chen2015microsoft}; each
image is paired with a corresponding sentence that describes the image (\ie, a
\emph{positive} example) and one that does not (\ie, a \emph{negative}
example). Negative sentences are collected by replacing object words in the
positive sentences with a similar object (\eg, changing the word ``dog'' to
``cat'' in ``The dog ran.'').  \citet{shekhar-etal-2017-foil} use the FOIL
dataset in a few tasks including a classification task where the model is asked
to classify if a sentence matches the image or not. We use the same task setup
because it allows us to probe image--language transformers in a zero-shot
setting as these models are generally trained to classify whether an
image--text pair match.
Our work is different than FOIL in that we focus on verb understanding as
opposed to noun understanding; moreover, our dataset provides different
negative types (by replacing subjects, verbs, or objects).

Other datasets focus on relationship or interaction detection \citep[\eg, HICO
and VRD;][]{chao:iccv2015,lu2016visual}.  These datasets are evaluated in a
classification setting in which the input is an image and the output is a
detected relationship (for HICO, an object and interaction, for VRD two objects
and their relationship) and have a limited number of verbs and objects.
V-COCO \citep{gupta2015visual} and ImSitu \citep{yatskar2016} both includes
verbs but do not provide negatives for a controlled evaluation of verb (or
noun) understanding.
Finally, other work has explored how creating hard negatives (\eg, by
substituting words in train examples) leads to better test performance
\citep[][]{gupta2020contrastive,hendricks2018grounding,faghri2017vse++}.  In
contrast, our work focuses on creating hard evaluation examples to probe
learned representations.

In summary, \datasetName{} is unique as it tests understanding of a broad range
of verbs as well as subjects and objects in a controlled way. Furthermore, our
dataset includes image--sentence pairs; thus, it can be used to evaluate
image--language transformers that process image--sentence pairs. Finally,
\datasetName{} is designed as a zero-shot task to evaluate pretrained
image--language transformers and is collected to have a similar distribution to
Conceptual Captions which is commonly used in pretraining these models. See
\Table{tab:datasets} for a comparison between \datasetName{} and other
datasets.

\begin{table}[]
\resizebox{\linewidth}{!}{
\begin{tabular}{l|rrrrcc}
dataset & Ims & Subjs &  Verbs & Objs & Sents & Negs \\ 
\hline
FOIL    & 32k & n/a &   0   & 70    &  \cmark  & \cmark  \\ 
HICO    & 10k & n/a & 117   & 80    & \xmark    & \cmark    \\ 
VRD    & 1k & 100 & 70   & 100    & \xmark      & \xmark \\ 
V-COCO    & 5k & n/a & 26   &  48  & \xmark     & \xmark   \\ 
ImSitu   & 25k & 950 & 504   & 1840    & \xmark  & \xmark      \\ 
\textbf{\datasetName{}} & 14k & 100 & 421 & 275 & \cmark & \cmark\\ 
\end{tabular}
}
\caption{\small \textbf{Im}ages, \textbf{Subj}ects, Verbs, \textbf{Obj}ects, \textbf{Sent}ences, and \textbf{Neg}atives in other datasets and \datasetName{}.  Image numbers are for the \textit{evaluation} set.}
\label{tab:datasets}
\end{table}

\ignore{
\begin{table}[]
\caption{Comparing HICO, FOIL, and \datasetName{}.  All numbers reported are for the \textit{evaluation} set.}
\resizebox{\linewidth}{!}{
\begin{tabular}{l|rrrrc}
dataset & Ims & Subjs &  Verbs & Objs & Sentences \\ 
\hline
HICO    & 10k & n/a & 117   & 80    & no        \\ 
FOIL    & 32k & n/a &   0   & 70    &  yes   \\ 
\datasetName{} & 14k & 100 & 421 & 275 & yes \\ 
\end{tabular}
}
\label{tab:datasets}
\end{table}
}

\section{Task Setup and Dataset Collection}

Our goal is to examine \emph{verb-understanding} in \emph{pretrained}
multimodal transformers.  To do so, we need a task that requires an
understanding of a given verb in a sentence, \eg, a model cannot succeed at the
task by relying on nouns. We also need to include a diverse set of verbs, and
examine each verb in at least a few situations.
To test the pretrained representations, we need to examine the models in a
zero-shot setting (without fine-tuning). 

Inspired by the FOIL setup \citep{shekhar-etal-2017-foil}, we use a zero-shot
classification task where a model is asked to identify if a sentence and an
image correspond to each other.
As a result, we need a dataset that provides ``match'' or ``not match'' labels
between images and sentences. 
We collect a dataset of image--sentence pairs (\datasetName{}) that given a
sentence, provides such labels for at least two images.\footnote{We note that
our dataset is limited to English sentences; we simply use ``sentences'' to
refer to English sentences.} Some of these images are \emph{positive} examples,
\ie, the sentence correctly describes them. Others are \emph{negative} examples
where some aspect of the sentence (\eg, verb) does not match the image. Figure
\ref{fig:qual-examples} shows some examples from our dataset.

We systematically collect negative examples such that they only differ from the
positive image with respect to the subject, verb, or object of the sentence. 
Finally, we consider sentences whose subjects, verbs, and objects are frequent
in the Conceptual Captions (CC) training dataset. Since CC is the dataset most
frequently used for pretraining multimodal transformers, we can examine what
the pretrained representations capture (in contrast to examining these models'
generalization ability).  We next describe our pipeline to create
\datasetName{}.

\paragraph{Creating a verb list.}  To ensure that we have a large number of
verbs in our dataset, we first created a verb list by considering a subset of
verbs that occur in the train split of the Conceptual Captions dataset
(CC-train).
More specifically, we consider verbs that are visually recognizable in the
images; to identify the visual verbs, we use the \emph{imSitu} dataset
\citep{yatskar2016} that includes verbs that annotators marked as reliably
recognizable. 
Moreover, we include verbs that occur at least 50 times in CC-train. 

\paragraph{Curating triplets.} 
Given a positive example, we need to systematically generate negatives by
replacing the subject, verb, or the object. As a result, we collect a set of
$\langle$subject, verb, object$\rangle$ (SVO) triplets from CC-train sentences
for our verbs.  We extract the subject, verb, and direct object from the
dependency parse trees and remove triplets where subjects or objects are
pronouns or have less than two characters.  Finally, we discard SVO triplets
with frequency smaller than five.

We consider three negative types for a given triplet: a subject-, verb-, or
object-negative where respectively, the subject, verb, or object in the triplet
are replaced by a different word. For example, given the triplet $\langle$girl,
lie, grass$\rangle$, examples of subject-negative, verb-negative, and
object-negative are $\langle$puppy, lie, grass$\rangle$, $\langle$girl, sit,
grass$\rangle$, and $\langle$girl, lie, beach$\rangle$.

Since our goal is to examine verb understanding, we only keep the triplets that
have at least one verb negative. This enables us to evaluate a model's capacity
in distinguishing images that mainly differ with respect to the verb; for
example, $\langle$girl, lie, grass$\rangle$ vs. $\langle$girl, sit,
grass$\rangle$.  Adding this constraint results in $11230$ SVO triplets and
$421$ verbs.
In this set, $1840$ SVO triplets (and $53$ verbs) have at least two verb and
object negatives.

\paragraph{Collecting images.} The next step is collecting images that match
the curated SVO triplets. We query for SVO triplets using the Google Image
Search API. We retrieve 5  images for each triplet, then remove any images with
urls in Conceptual Captions.
To make sure that these automatically-retrieved images certainly match the
triplets, we set up an annotation task where we ask workers on Amazon
Mechanical Turk (AMT) to verify if the subject, verb, and object are present in
the image. 
We ask three people to annotate each image, and only keep images where at least
two annotators agree that the subject, verb, and object are depicted in the
image. Moreover, we discard images marked as a cartoon by annotators.  We find
that 58\% of our images pass this initial annotation process.  We pay workers
\$0.04 per HIT for all tasks.

\paragraph{Collecting sentences.}
Multimodal transformer models are trained on pairs of images and
\textit{sentences}; to evaluate them, we require image--sentence pairs as
opposed to image--SVO pairs.  Given an image and an SVO triplet, we next ask
annotators to write a sentence that uses all the words in the triplet and
describes the image.  
For example, as shown in Figure \ref{fig:qual-examples} top right, given the
triplet $\langle$man, jump, sea$\rangle$, an annotator might write ``A man is
jumping into the sea.''.
We ask annotators to refrain from writing additional information to ensure that
a collected sentence examines the words in the SVO (as opposed to words that we
are not controlling for).
Annotators are given the option to not write a sentence if they do not think
the subject, verb, and object can be combined into a grammatical sentence that
describes the image.
86\% of our images pass this phase of our pipeline.

We observe that for a given SVO, different images elicit slightly different
sentences.  For example, the triplet $\langle$person, jog, beach$\rangle$
resulted in the sentences ``A person jogging along the beach.'' and ``A person
jogs at the beach.''.  Additionally, annotators pluralize nouns to ensure the
sentence describes the image (\eg, Figure \ref{fig:qual-examples} top left, the
subject ``child'' is written as ``children'' in the sentence).

\paragraph{Confirming the negative image.} Finally, given a \emph{positive}
triplet (\eg, $\langle$girl, lie, grass$\rangle$) and its negative (\eg,
$\langle$girl, sit, grass$\rangle$), we need to confirm that the positive's
sentence does not match the image retrieved for the negative triplet. To do so,
we ask three annotators to select which images (positive, negative, neither, or
both) match a given sentence.  Image--sentence pairs where two out of three
annotators agree are accepted into our dataset; 68\% of the pairs pass this
final annotation stage.

\begin{table*}[]
\resizebox{\linewidth}{!}{
\begin{tabular}{@{}l|ll|cc|l@{}}
\toprule
 Name    & Multimodal Attention & Similar Model & MLM & MRM &  \zsflickr \\ \midrule
  \mmt       & Queries from L (I) take values and keys from \emph{only} I (L) & \vilbert; LXMERT        &  \cmark       &   \cmark & \textbf{41.9} \\
  \mmtSS     & Queries from L (I) take values and keys from \emph{both} L and I  &    UNITER            &  \cmark    &   \cmark & \textbf{40.0} \\ 
  \mmtLQ     & Queries are \emph{only} from L  \citep[][]{hendricks2021decoupling}  &     &  \cmark     &  \cmark & 33.6\\ 
  \mmtIQ     & Queries are \emph{only} from I  \citep[][]{hendricks2021decoupling}  &          &  \cmark     & \cmark &  31.6\\ 
  \smt       & Single-Modality Transformers without multimodal attention   &       &  \cmark     & \cmark &  16.9\\ 
  \hline
  \mmtNoMRM       & The same as \mmt &            &  \cmark    & \xmark &  \textbf{41.1}\\ 
  \mmtNoMLM       & The same as \mmt  &           &  \xmark    & \cmark & 20.2 \\ 

\bottomrule
\end{tabular}
}
\caption{Different variants of the image--language transformer architecture we test. L and I stand for language and image, respectively.
 We note that models with Merged attention (like UNITER) are also referred to as single-stream models.
\vilbert: \citet{lu2019vilbert}; LXMERT: \citet{tan2019lxmert}; UNITER: \citet[][]{chen2019uniter}}
\label{tab:models}
\end{table*}

\section{Experimental Setup and Results}
We investigate if current image--language transformers can relate different
aspects of language (and in particular verbs) to images by evaluating these
models against both FOIL and \datasetName{}. More specifically, we evaluate a
few architectural variations of image--language transformer models \citep[based
on the implementation of the models by][]{hendricks2021decoupling} that differ
in their choice of multimodal attention and loss functions; this way we can
examine whether our findings are sensitive to these slight differences.
The base multimodal transformer (\textbf{\mmt}) closely replicates the \vilbert
architecture \citep{lu2019vilbert}: this model includes three loss functions,
masked language modeling (MLM) and masked region modeling (MRM) losses on the
language and image inputs and an image--text matching (ITM) loss that
classifies if an image--sentence pair match. 
Importantly, the multimodal attention of \mmt is similar to the hierarchical
co-attention in \citet{lu2016hierarchical} where each modality (\ie, image or
language) attends \emph{only} to the other modality. More specifically, in the
multimodal self-attention layer of transformer \citep{vaswani2017attention},
for queries on the language input, keys and values are taken from images and
vice versa.

Different interactions of image (language) queries, keys, and values in
multimodal self-attention results in variations of image--language
transformers. 
We describe the model variations we study in \Table{tab:models}.  We also
consider models that either lack the MLM or MRM loss.  Models are pretrained on
Conceptual Captions (CC) unless stated otherwise.  For reference, we report the
Recall@1 performance on the zero-shot image-retrieval task on Flickr
(\zsflickr), where a model must retrieve an image from the Flickr dataset
\cite{young2014image} that matches an input sentence.  Since \mmt performs best
on \zsflickr, we do most of our experiments on this model unless stated
otherwise.

We first evaluate our image--language transformers on FOIL to examine their
noun understanding and then test them on  \datasetName{} which probes for
subject, verb, and object understanding in learned representations.
Following FOIL, we report the accuracy on positive and negative pairs.
All our models have an image-text classification output used in pretraining to
align images and sentences.  We calculate accuracy by passing images through
our models and labeling an image--sentence pair as negative if the classifier
output is $<0.5$ and positive otherwise.  We report the average over the two
pairs (see Avg columns in Tables \ref{tab:foil} and \ref{tab:accuracy_by_type})
by weighting them equally, since we expect models to perform well on both
positive and negative pairs. In FOIL, there are equal positive and negative
pairs.

Another possible way to set-up our evaluations is as image-retrieval (reporting
recall@1 as a metric). However, the retrieval setting does not highlight the
difference in performance between positive and negative pairs. For example, a
model might rank the pairs correctly even when their scores are very close
(positive score is 0.91 and negative one is 0.9). In this example, the model is
wrong about the negative pair (it is assigned a high score) but the retrieval
setting does not capture this. However, the classification metric will penalize
the model for assigning a high score to a negative pair. As a result, the
classification metric better differentiates between the models by examining if
they correctly label both the positive and negative pairs.

\subsection{Evaluating Nouns with FOIL}  We examine noun understanding in
image--language transformers with the FOIL dataset
\citep{shekhar-etal-2017-foil}. 
Given image--sentence pairs from FOIL, we evaluate the \mmt model in a
zero-shot setting by using it to classify if the image and sentence match.
\Table{tab:foil} compares \mmt with the best model from the FOIL paper
\citep[HieCoAtt][]{shekhar-etal-2017-foil} and, to our knowledge, the
best-performing model on the task without using ground-truth annotations
\citep[Freq+MM-LSTM from][]{madhyastha2018defoiling}. 
Note that these models are trained specifically for the FOIL task (\ie, on the
train split of FOIL), whereas the \mmt model (pretrained on CC) is tested in a
zero-shot setting.

\mmt achieves an accuracy considerably worse than the best models on FOIL
\citep{shekhar-etal-2017-foil,madhyastha2018defoiling} on all pairs; this is
surprising given that image--language transformers achieve state-of-the-art
results on zero-shot image retrieval tasks based on Flickr
\citep{young2014image} and MSCOCO \citep{chen2015microsoft}. In particular,
\mmt overpredicts that image--sentence pairs match, resulting in the highest
accuracy on the positive pairs ($99.0$) but the lowest on negative pairs
($11.8$). Thus \mmt cannot distinguish between sentences that only differ with
respect to nouns.
We investigate whether this poor performance of \mmt is due to mismatch between
the pretraining (\ie, CC) and FOIL test (\ie, MSCOCO) datasets.
Thus, we compare our \mmt model pretrained on Conceptual Captions with one
pretrained on MSCOCO (MMT-COCO). As expected, MMT-COCO has considerably higher
performance on all pairs (compare to \mmt);  however, the accuracy is still
significantly higher on positive pairs than negative ones, showing that the
model overpredicts that image--sentence pairs match.
Our result shows that despite their impressive performance on downstream tasks,
image--language transformer models perform poorly in distinguishing between
semantically similar sentences. Next we examine how well these models perform
on our proposed probing dataset which is designed to have a similar vocabulary
to the CC pretraining dataset.

\begin{table}
    \centering
    \begin{tabular}{l|rrr}
    Model & Avg & Pos. & Neg. \\
    \midrule
    HieCoAtt*                & 64.1 & 91.9 & 36.4 \\
    Freq + MM-LSTM $\dagger$ & \textbf{87.9} & 86.7 & \textbf{89.0} \\
    \midrule
    \mmt                     & 55.4 & \textbf{99.0} & 11.8 \\
    MMT-COCO                 & 72.0 & 95.0 & 49.0 \\
    \end{tabular}
    \caption{Performance on FOIL averaged over all (Avg), positive (Pos.), and negative (Neg.) pairs. *\citet{shekhar-etal-2017-foil}; $\dagger$\citet{madhyastha2018defoiling}.
    }
    \label{tab:foil}
\end{table}

\begin{table*}
\centering
%
\resizebox{\linewidth}{!}{
\begin{tabular}[h]{ l|rrr|rrr|rrr|rrr } 
 \hline
 & \multicolumn{3}{c}{Overall} & \multicolumn{3}{|c|}{Subj. Negative} & \multicolumn{3}{|c|}{Verb Negative} & \multicolumn{3}{|c}{Obj. Negative}  \\
   & \multicolumn{1}{c}{Avg} & \multicolumn{1}{c}{Pos.} & \multicolumn{1}{c|}{Neg.} &  \multicolumn{1}{c}{Avg} & \multicolumn{1}{c}{Pos.} & \multicolumn{1}{c|}{Neg.} & \multicolumn{1}{c}{Avg} & \multicolumn{1}{c}{Pos.} & \multicolumn{1}{c|}{Neg.} & \multicolumn{1}{c}{Avg} & \multicolumn{1}{c}{Pos.} & \multicolumn{1}{c}{Neg.} \\
  \# Examples & 48k & 12k & 36k &   8k & 3k & 5k &   34k & 11k & 23k &  11k & 3k & 8k \\

 \midrule
  \mmt & 64.3 & \textit{93}.8 & 34.8 &
67.0 & \textit{94.4} & 39.5 &
60.8 & \textit{93.8} & 27.8 &
73.4 & \textit{94.4} & 52.4 \\
  \mmtSS & 64.7 & \textbf{94.4} & 35.0 &
69.1 & \textbf{94.9} & 43.2 &
60.7 & \textbf{94.4} & 27.0 &
74.1 & \textbf{94.9} & 53.3 \\
  \mmtLQ & \textit{68.1} & 80.2 & \textbf{56.0} &
\textit{71.5} & 82.1 & \textbf{60.9} &
\textit{64.5} & 80.2 & \textit{48.9} &
\textit{77.7} & 81.4 & \textbf{74.1} \\
  \mmtIQ & 64.3 & 91.6 & 37.0 &
68.2 & 92.1 & 44.2 &
59.7 & 91.6 & 27.8 &
75.6 & 91.5 & 59.6 \\
  \smt & 52.4 & 49.1 & \textit{55.6} &
52.6 & 47.7 & 57.5 &
51.8 & 49.1 & \textbf{54.6} &
53.9 & 50.7 & 57.0 \\
   \hdashline
\mmtNoMRM & \textbf{69.5} & 85.4 & 53.7 &
\textbf{73.5} & 87.4 & \textit{59.7} &
\textbf{65.5} & 85.6 & 45.5 &
\textbf{80.1} & 86.2 & \textbf{74.1} \\
\mmtNoMLM & 60.8 & 92.3 & 29.3 &
64.8 & 93.9 & 35.8 &
57.4 & 92.5 & 22.4 &
69.5 & 93.6 & 45.5 \\
 \hline
\end{tabular}
}
\caption{Results on \datasetName{} on different models for subject, verb, and object negatives.  Best results are shown in bold; second best results are italicized.}
\label{tab:accuracy_by_type}
\end{table*}

\subsection{Comparing Models on \datasetName{}}  
\label{sec:overall_results}

We evaluate all models (see \Table{tab:models}) on \datasetName{} and report
overall accuracy and accuracy for subject, verb, and object negatives in
\Table{tab:accuracy_by_type}.

The \mmt model (with the best performance on \zsflickr) performs poorly on
\datasetName{},  achieving an overall average accuracy of $64.3$.
The best overall average accuracy (\mmtNoMRM; $69.5$) shows that \datasetName{}
is challenging for image--language transformers. In particular, models struggle
with classifying negative pairs; \mmtLQ achieves the highest accuracy over
negative pairs ($56$) which is slightly higher than chance at $50$.
\footnote{We focus on image--language transformers, but we also tested a
baseline model where image features are embedded with the detector used in our
transformers and language features with BERT.  Features are pooled using
element-wise multiplication.  This baseline achieves 66.3\% accuracy overall
with 75.4\% and 57.3\% accuracy on positives and negatives.  Similar to
transformers, performance on verbs is the worst.}

Though \mmtNoMRM and \mmt perform similarly on \zsflickr, \mmtNoMRM performs
better on \datasetName{}.  This suggests that the masked region modelling loss
is not needed for good performance on \zsflickr; also, it impedes the model
from learning fine-grained representations needed to perform well on
\datasetName{}.
More surprisingly, \mmtLQ, which performs worse on \zsflickr than \mmt,
outperforms \mmt on \datasetName{}.  The image representations in \mmtLQ are
not updated with an attention mechanism.  In \Section{sec:image_sim}, we
explore if the stronger attention mechanism in \mmt leads to overfitting of the
training images and thus weaker performance.

We crafted \datasetName{} such that it includes words from the pretraining
dataset of image--language transformers (\ie, CC), whereas FOIL is collected
from MSCOCO.  Comparing the performance of \mmt (with CC pretraining) on FOIL
and \datasetName{} ($55.4$ in \Table{tab:foil} vs. $64.3$ in
\Table{tab:accuracy_by_type}), we see that the domain mismatch between
pretraining and test data plays a role in \mmt's performance.
Interestingly, comparing the performance of \mmt-COCO (\mmt with COCO
pretraining) on FOIL to \mmt (with CC pretraining) on \datasetName{}, we find
that \datasetName{} is more challenging than FOIL when there is no domain
mismatch ($72.0$ in \Table{tab:foil} vs. $64.3$ in
\Table{tab:accuracy_by_type}).

When comparing different negative types across \emph{all} models, we observe
that verbs are harder than subjects and objects; compare average accuracy for
Subj., Verb, and Obj. Negative columns in \Table{tab:accuracy_by_type}. For
example, in \mmt, the subject and object negative average accuracies ($67.0$
and $73.4$) are considerably higher than the average accuracy for verb
negatives ($60.8$).
Moreover, when breaking down the accuracies for positive and negative pairs
({Pos. and Neg. columns in \Table{tab:accuracy_by_type}}), we observe that the
accuracies of positive pairs are similar (ranging between $80.2$ and $94.4$)
across all models except \smt (which performs close to chance); however, for
negative pairs, there is more variation in accuracy across models especially
for verb negatives (ranging between $22.4$ and $54.6$, {Neg. columns under
``Verb Negative''}).
These results show that negative pairs are better than positive ones in
distinguishing between different model architectures.
%

We also find that subjects are harder than objects across all models (when
comparing average accuracies of subject and object negatives).  To better
understand this result, we examined $21$ nouns that occur both as subjects and
objects in \datasetName{}' sentences. Interestingly, over these $21$ nouns, for
our \mmt model, the accuracies of negative pairs are $42.9$ and $56.4$ for
subject and object negatives, respectively. This suggests that the subject
position might be more challenging than the object one which we further explore
in \Section{sec:freq_and_acc}.

\subsection{Accuracy and Frequency at Training} 
\label{sec:freq_and_acc}

\begin{figure}[t!]
        \centering
        \begin{subfigure}[]{.45\textwidth}
            \centering
            \includegraphics[width=.9\linewidth]{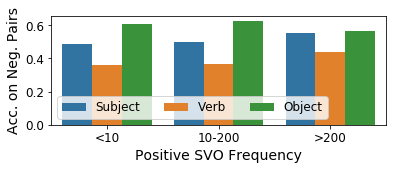} 
            \label{fig:freq-pos}
        \end{subfigure}
        \begin{subfigure}[]{.45\textwidth}
            \centering
            \includegraphics[width=.9\linewidth]{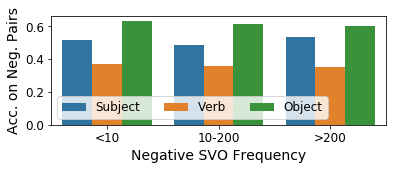} 
            \label{fig:freq-neg}
        \end{subfigure}
        \caption{Accuracy of negative pairs for subject, verb, and object negatives given SVO frequencies in CC.}
        \label{fig:freq-plots} 
\end{figure}

Our overall results on \datasetName{ } (\Table{tab:accuracy_by_type}) show that
for image--language transformers, verb negatives are more challenging  than
subject and object ones, and also subject negatives are harder than object
ones.
We examine if this observation is due to properties of \datasetName{} as
opposed to differences specific to subjects, verbs, and objects.
First, we explore whether the frequency of SVO triplets in pretraining data
impacts the accuracy of negative pairs in our \mmt model. We focus on negative
pairs as there is more variation in negative-pair accuracies across both models
as well as subject, verb, and object negatives. We consider the frequency of
positive and negative SVO triplets: a positive SVO corresponds to a positive
image matching a given sentence, but a negative SVO and its extracted negative
image do not match the sentence.

We group SVOs based on their frequency in CC-train into low (less than 10),
medium (between 10-200), and high (greater than 200) frequency bins.
\Figure{fig:freq-plots} plots the negative-pair accuracy for subject, verb, and
objects across these different frequency bins over positive and negative SVO
frequencies.
We confirm that our result on the difficulty of negative types does not depend
on the frequency of positive or negative SVOs in pretraining data. In both
plots of \Figure{fig:freq-plots}, the negative types in order of difficulty
(lower accuracy) are verbs, subjects, and objects independent of the frequency
bin.

\paragraph{Similarity between SVOs.}  
We examine if the similarity between the SVO triplets corresponding to the
negative and positive images can explain the difference in performance of
subject, verb, and object negatives.
For example, we expect that distinguishing  $\langle$child, cross,
street$\rangle$ and $\langle$adult, cross, street$\rangle$ to be harder than
differentiating one of them from $\langle$dog, cross, street$\rangle$:
``child'' and ``adult'' are more similar to each other than to ``dog''.  To
test this, we measure the similarity between subjects, verbs, and objects in
their corresponding negative types using the cosine similarity between word2vec
\cite{mikolov2013distributed} embeddings.

The average similarities between subjects, verbs, and objects are $0.49$,
$0.29$, $0.27$, respectively.  Thus, subject words in negative examples tend to
be more similar than object words.  Furthermore, we find that there is a small
positive correlation (as measured by Spearman rank correlation) between SVO
similarity and classifier scores for negative pairs -- $.264$ and $.277$ for
subjects and objects respectively -- suggesting that when SVOs corresponding to
the image and sentence are similar, the classifier tends to assign a higher
score (more positive) to the pair.  This partially explains why accuracy on
subjects is lower than on objects in Table \ref{tab:accuracy_by_type}.
Even though verb negatives are harder for our model, the similarity for verb
negatives is similar to that of object negatives.  The correlation coefficient
between similarity and classifier score is weaker ($.145$) for verbs,
suggesting that word similarity factors less in how well the model classifies
verb negatives.

\subsection{Similarity to Pretraining Data}
\label{sec:image_sim}

\begin{figure}[]
\centering
\includegraphics[width=0.7\linewidth]{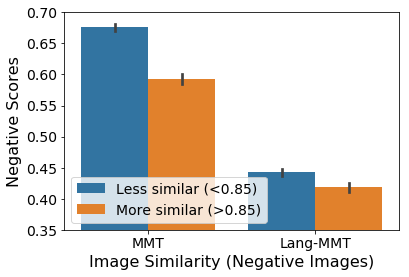}
\caption{Comparing negative scores on \mmt and \mmtLQ for images less or more similar to CC.}
\label{fig:image-sim}
\end{figure}

We next consider the similarity between images in \datasetName{} and CC.  To
measure the similarity between images, we sample 1 million images from CC.  For
each image in \datasetName{}, we find the 10 nearest neighbors in the feature
embedding space of CC, and average the distance to compute a similarity score
for the image. 
Figure \ref{fig:image-sim} plots the average score from our classifier for
negative pairs with images that are less or more similar to the pretraining
data (since we are classifying negative pairs, the lower score the better).  We
compare the \mmt and \mmtLQ models since they have considerably different
performance on \datasetName{} and \zsflickr. The difference in average scores
between less similar and more similar examples for \mmt is $0.083$.  This is
noticeably greater than the difference in average scores between less and more
similar examples for \mmtLQ ($0.024$), suggesting that the image similarity
influences \mmtLQ less than \mmt.  One hypothesis is that the stronger
attention mechanism in \mmt overfits to the training images which makes the
\mmt model less robust.

\subsection{The Choice of Pretraining Dataset}

In \Section{sec:overall_results}, we observe that models perform particularly
poorly in classifying negative pairs. We investigate whether the choice of
pretraining dataset impacts this observation. Conceptual Captions (CC), the
most-common pretraining dataset for image--language transformers, is curated by
scraping images and alt-text captions from the web. As a result, compared to
manually-annotated datasets such as MSCOCO, CC is noisy -- it contains examples
where the sentence and its corresponding image do not completely align. For
example, a sentence can mention objects that are not in the image or, in
extreme cases, does not describe the image at all.

We hypothesize that image--language transformers treat correspondences due to
dataset noise as ``real'' relations; in other words, they learn that if a
image--sentence pair is somewhat semantically related, it should be classified
as a positive match, even if some aspects of the sentence do not describe the
image.
At the same time, we can think of negatives in \datasetName{} as examples with
noisy correspondences where a specific aspect of a sentence (\eg, the verb)
does not match the image. 
We compare our \mmt model (with CC pretraining) to one trained on a
manually-annotated and less noisy dataset, MSCOCO (referred to as MMT-COCO). 

\Table{tab:datasets-eval} reports the overall accuracy of the two models on
\datasetName{} as well as a breakdown over subject, verb, and object negatives
for negative-pair accuracies.
MMT-COCO performs better than \mmt pretrained on CC (avg.\ accuracy of $68$ vs
$64.3$). This is surprising since MMT-COCO has a different image and language
distribution in its pretraining dataset.
The accuracy of positive pairs in MMT-COCO is considerably lower than \mmt
while it performs noticeably better for negative pairs: unlike \mmt, the
MMT-COCO model does not overpredict that image--sentence pairs match.
Our results show the image--language transformers are not robust to dataset
noise. 
Less-noisy datasets (such as MSCOCO), despite their small size and domain
mismatch, are more suitable for learning representations that are sensitive to
finer-grained differences in images.
Alternatively, models which are more robust to noise in datasets could be
beneficial for tasks like ours.

\begin{table}
    \centering
    \resizebox{\linewidth}{!}{
    \begin{tabular}{l|rrr|rrr}
    & \multicolumn{3}{c|}{Overall} & \multicolumn{3}{c}{Neg.  Acc.} \\
    Train & \multicolumn{1}{c}{Avg.} & \multicolumn{1}{c}{Pos.} & \multicolumn{1}{c|}{Neg.} & \multicolumn{1}{c}{S} & \multicolumn{1}{c}{V} & \multicolumn{1}{c}{O} \\
    \midrule
    CC & 64.3 & 93.8 & 34.8 &39.5 & 27.8 & 52.4 \\
    COCO & 68.0 & 75.2 & 60.9 & 66.0 & 55.5 & 73.4\\
    \end{tabular}
    }
    \caption{Comparing performance when training our MMT model on COCO and CC.}
    \label{tab:datasets-eval}
\end{table}

\subsection{Which Verbs Are the Hardest?}  
We investigate which verbs are hardest for \mmt.
We consider verbs with many examples in \datasetName{}: we keep SVO triplets
with at least 30 negative images, resulting in a set of $147$ verbs and $887$
SVO triplets across $4,843$ images.
Table \ref{tab:hard-verbs} lists the easiest and hardest verbs (with highest
and lowest accuracy for negative pairs) for the \mmt model.
Easy and hard verbs have a diverse set of properties; for example, easy verbs
include sporting activities like ``tackle'' as well as verbs like ``lead'' that
occurs in a variety of contexts.  We also examine the 20 most difficult and
easiest verbs for \emph{all} our models (described in \Table{tab:models}).
Most difficult verbs for all models include: ``cut'', ``argue'', and``break''
and the easiest ones include: ``direct'', ``battle'', ``surround'', ``skate'',
and ``participate''.

We test if verbs that occur in both \datasetName{} and imSitu are easier for
our model to classify.  Verbs in imSitu are considered visual as the dataset
collection pipeline for imSitu includes an explicit annotation step to
determine if verbs are visual.  Surprisingly, we find verbs in imSitu are
harder for our \mmt model.  On closer inspection, some verbs in our dataset but
\emph{not} in imSitu (\eg, ``swim'') are clearly visual.
An interesting future direction is to investigate which visual properties of a
verb make it harder or easier for image--language models to learn.

\begin{table}
    \centering
    \resizebox{\linewidth}{!}{
    \begin{tabular}{c|c}
    Easy & Hard \\
    \midrule
    \small{tackle, reach, arrive, pitch,} & \small{argue, beat, break,} \\
    \small{accept, congratulate, lead,} & \small{burn, buy, cast, comb,} \\
    \small{present, celebrate, attend} & \small{crash, cut, decorate} \\

    \end{tabular}
    }
    \caption{Hard and easy verbs for our MMT model}
    \label{tab:hard-verbs}
\end{table}

\section{Conclusions}

Although image--language transformers achieve impressive results on downstream
tasks, previous work suggests performance on these tasks can be confounded by
factors such as over-reliance on language priors \citep{goyal2017making}.  We
collect a dataset of image--sentence pairs to examine  multimodal understanding
by testing the ability of models to distinguish images that differ with respect
to subjects, verbs, and objects. 

Our results show that image--language transformers fail at identifying such
fine-grained differences; they incorrectly classify image--sentence pairs that
do not match.  Surprisingly, a model trained on a manually-annotated and
smaller dataset does better on our task, suggesting that models have trouble
ignoring noise in larger but automatically-curated pretraining datasets.
Additionally, verb understanding is harder than subject or object understanding
across all models we study.  This motivates the need for researchers to not
only examines models on objects, but develop datasets and architectures which
allow for better verb understanding as well.

\section*{Acknowledgements}
We would like to thank Antoine Miech for detailed comments on our paper.
   Also, thanks to Phil Blunsom, Laura Rimell, Stephen Clark, Andrew Zisserman,
   Jean-Baptiste Alayrac and the anonymous reviewers for their helpful
   feedback. We also thank Susie Young, Zhitao Gong, and Tyler Liechty for
   supporting our data collection.


\bibliographystyle{acl_natbib}
\bibliography{references}

\end{document}